%% file: main.tex
\documentclass[10pt,twocolumn,letterpaper]{article}

\usepackage[cvprfinal]{cvpr}      


\definecolor{cvprblue}{rgb}{0.21,0.49,0.74}
\usepackage[pagebackref,breaklinks,colorlinks,allcolors=cvprblue]{hyperref}
\usepackage{multirow} 
\usepackage{makecell}

\def\paperID{*****} 
\def\confName{CVPR}
\def\confYear{2025}

\title{Learning to Drive from a World Model}

\author{Mitchell Goff\qquad Greg Hogan\qquad George Hotz\qquad Armand du Parc Locmaria\qquad Kacper Raczy\\
Harald Schäfer\qquad Adeeb Shihadeh\qquad Weixing Zhang\qquad Yassine Yousfi\\
\textbf{comma.ai}\\
{\tt\small{autonomy@comma.ai}}
}

\begin{document}
\maketitle
\input{0_abstract}
\input{1_introduction}

\input{9_formulation}
\input{9_reprojective_simulation}
\input{9_worldmodel_simulation}
\input{9_driving_policy_training}
\input{9_conclusion}

{
    \small
    \bibliographystyle{ieeenat_fullname}
    \bibliography{main}
}


\end{document}

%% file: 0_abstract.tex
\begin{abstract}

Most self-driving systems rely on hand-coded perception outputs and engineered driving rules. Learning directly from human driving data with an end-to-end method can allow for a training architecture that is simpler and scales well with compute and data.

In this work, we propose an end-to-end training architecture that uses real driving data to train a driving policy in an on-policy simulator. We show two different methods of simulation, one with reprojective simulation and one with a learned world model. We show that both methods can be used to train a policy that learns driving behavior without any hand-coded driving rules. We evaluate the performance of these policies in a closed-loop simulation and when deployed in a real-world advanced driver-assistance system.

\end{abstract}

%% file: 1_introduction.tex
\section{Introduction}
\label{sec:introduction}

Autonomous driving has seen remarkable progress in recent years, with learning-based approaches replacing increasing portions of the system. However, despite these advancements, most self-driving products still rely on handcrafted rules on top of a layer of perception. These modular methods require significant engineering effort to generalize to real-world complexities and edge cases. Instead, End-to-End (E2E) learning offers a more scalable solution by training a driving policy to imitate human driving behavior from real data. An E2E policy can take in raw sensor inputs, such as images, and directly output a driving plan or control action, eliminating the need for manual rule design \cite{bojarski2016end}.

A key challenge in E2E learning is how to train a policy that can perform well under the non-i.i.d. assumption made by most supervised learning algorithms such as Behavior Cloning \cite{bain1995framework}. In the real world, the policy's predictions influence its future observations. Small errors accumulate over time, leading to a compounding effect that drives the system into states it never encountered during training.

To overcome this, the driving policy needs to be trained on-policy, allowing it to learn from its own interactions with the environment, and enabling it to recover from its own mistakes. Running on-policy learning in the real world is costly and impractical \cite{kendall2019learning}, making simulation-based training essential.

Traditional driving simulators are often handcrafted with explicit limited traffic behaviors and scenes. While useful for testing, these simulators fail to capture the full complexity and richness of real-world driving.

In this work, we explore how two data driven simulators can be used to train an E2E driving policy: a reprojective novel view synthesis simulator \cite{seitz1996view}, and a learned World Model \cite{ha2018worldmodels,santana2016learningdrivingsimulator, hu2023gaia1generativeworldmodel}. By using real-world data these simulators can capture the full diversity of real-world driving scenarios, and provide ground-truth for policy decisions by imitating the human driving decisions. We propose a method to distill human driving behaviors during on-policy training, by anchoring a supervising model to future states.

We discuss limitations of the reprojective simulator, and the potential of the World Model simulator to scale with data and compute to overcome these limitations. We show that policies trained in this way learn normal driving behaviors, such as staying in a lane and changing lanes, and that they can be used in real-world applications, such as an Advanced Driver Assistance System (ADAS).

To our knowledge, this is the first work to show how end-to-end training, without handcrafted features, can be used in a real-world ADAS. Additionally, we believe this is the first use of a world model simulator for on-policy training of a policy that is deployed in the real world.

\input{world_model_block_diagram.tex}

%% file: world_model_block_diagram.tex
\begin{figure}[ht]
    \centering
    \includegraphics[width=0.99\linewidth]{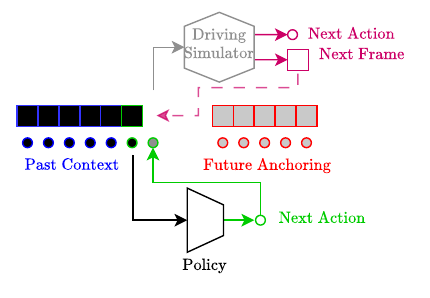}
    \caption{One step of the World Model Simulation rollout. Gray filled shapes are inputs to the World Model. Black filled shapes are inputs to both the Policy Model and the World Model (note that the Policy Model can be the World Model itself). Circles are actions (positions and orientations) and rectangles are observations (images).}
    \vspace{-0.05in}
    \label{fig:world_model_block_diagram}
\end{figure}

%% file: 9_formulation.tex
\section{Formulation}
\label{sec:formulation}

\subsection{Driving Policy}
\label{sub:driving_policy}

Our goal is to learn an End-to-End (E2E) driving policy $\pi$ that maps from a history of observations $(o_1, o_2, \ldots, o_T)$ to a distribution over next actions $a_{T+1}$. We consider a history $h_T^{\pi}$  to be a sequence of observations and previous actions.

\begin{equation}
\label{eq:driving_policy}
    \begin{split}
        \pi: h_T^{\pi} &\mapsto p(a_{T+1} \mid h_T^{\pi}), \\
        \text{with } h_T^{\pi} &= \Bigl((o_1, a_1), (o_2, a_2), \ldots, (o_T, a_T)\Bigr).
    \end{split}
\end{equation}

The action space $\mathcal{A}$ is defined as a desired turning curvature and a desired longitudinal acceleration. The observation space $\mathcal{O}$ is defined as a set of camera images only (Vision Only Policy). For simplicity, we will only show images coming from a single camera (narrow field of view). In practice, we use images from two cameras: wide and narrow field of view in order to capture a larger portion of the scene. The formulation can be extended to include more cameras without loss of generality.

We are given a dataset of expert demonstrations $\mathcal{D} = \Bigl\{\Bigl((s_1, a_1), \ldots, (s_T, a_T)\Bigr)\Bigr\}_{i=1}^n$. We aim to learn a driving policy $\pi$ given the expert demonstrations in $\mathcal{D}$.

The state space $\mathcal{S}$ is defined as the set of camera images, but can also include other sensor data such as GPS, and IMU data. In this work, we restrict the state space to the set of camera images $\mathcal{O}$ and a global pose estimate (position and orientation) of the vehicle $p_t=(x,y,z,\phi,\theta,\psi) \in \mathcal{P} \subset \mathbb{R}^6$.

The global pose is obtained using a tightly coupled GPS/Vision Multi-State Constraint Kalman Filter (MSCKF) \cite{schafer2018commutedatacomma2k19dataset,mourikis2007msckf, mingyang2013highprecisionekf}.

\subsection{Driving Simulator}
\label{sub:driving_simulator}

We define a Driving Simulator as composed of (i) a Driving State Generator and (ii) an Action Ground Truth Source.

The Driving State Generation can be based on traditional driving simulators such as CARLA \cite{Dosovitskiy17}, MetaDrive \cite{li2021metadrive}, etc. or based on a so-called World Model that is learned from data (Section \ref{sec:world_models_simulation}), or based on reprojective novel view synthesis techniques (Section \ref{sec:reprojective_simulation}).

The Action Ground Truth Source provides the supervision signal for training the driving policy. We describe a data-driven Action Ground Truth Source in Section \ref{sub:driving_ground_truth}.

\subsection{Vehicle Model}
\label{sub:vehicle_model}

A vehicle's motion is described as a sequence of poses $(p_1, p_2, \ldots, p_T)$. To simulate the effects of actions taken in simulation, a function is needed that produces poses based on actions. We call this a Vehicle Model \cite{jazar2008vehicle}. Our Vehicle Model is designed to model a variety of real-world effects including vehicle dynamics, delayed steering response, wind, and more. Simulating these effects is needed for transferring policies trained in simulation to the real world, and is often referred to as Domain Randomization (sim2real) \cite{sadeghi2017cad2rl, tobin2017domainrandomization, peng2018simtoreal}. By inverting the Vehicle Model we can also estimate actions needed to achieve a trajectory of poses.

Figure \ref{fig:world_model_block_diagram} illustrates the different building blocks involved in one step of a driving simulator.

\subsection{World Model}
\label{sub:world_model}

We define a World Model $w$ as a stochastic model that predicts a future state given a history of states and actions. In order to make the World Model independent of the Vehicle Model described in \ref{sub:vehicle_model}, we consider the actions (desired curvature and acceleration) to be implicitly deducible from the vehicle's poses $(p_1, p_2, \ldots, p_T)$ and the Vehicle Model. In other words, $w$ maps a history of images and poses and next pose to a distribution of the next image.

\begin{equation}
\label{eq:world_model}
    \begin{split}
        w: h_T^w &\mapsto p(o_T \mid h_T^w), \\
        \text{with } h_T^w &= \Bigl((p_1, o_1), (p_2, o_2) \ldots, (p_T,)\Bigr).
    \end{split}
\end{equation}

Note that using the pose as the transition signal for the World Model enables augmenting the Vehicle Model's parameters without needing to retrain the World Model.

\subsection{Future Anchored World Model}
\label{sub:future_anchored_world_model}

We can train non-causal World Models similar to \cite{baker2022video} conditioned on future observations and actions parametrized by $F=(f_s, f_e)$, where $f_s$ is the start of the future horizon and $f_e$ is the end of the future horizon. With $f_s>T$. This model can only be used offline, but has the advantage of predicting human-like driving video sequences and trajectories that converge to a goal state at $F$. We refer to this as recovery pressure.

\begin{equation}
\label{eq:future_anchored_world_model}
        \begin{split}
            w: h_{T,F}^w &\mapsto p(o_T \mid h_{T,F}^w), \\
            \text{with } h_{T,F}^w &= \Bigl((p_{f_s}, o_{f_s}), \ldots, (p_{f_e}, o_{f_e}), \\
            &\quad (p_1, o_1),(p_2, o_2), \ldots, (p_T,)\Bigr).
        \end{split}
\end{equation}

\subsection{Driving Ground Truth}
\label{sub:driving_ground_truth}

The Action Ground Truth at time $T$ refers to the actions $a_T \mid h_T$ the policy should take, given the past observations and the past actions it has taken, resulting in good driving behavior.

To generate this ground truth, we train the Future Anchored World Model to also predict the next pose or trajectory of poses $\mathcal{T}$ given a history of observations and poses and a Future Anchoring, $\mathcal{T} \mid h_{T,F}$. When running in this mode, the World Model is referred to as a Plan Model, and the predicted trajectory can be mapped to ground-truth actions using the Vehicle Model.

Future Anchoring is essential for enabling the Plan Model to produce a trajectory that converges to a desirable goal state, $F$, regardless of the current state of the simulation, without it, the Plan Model does not exhibit recovery pressure when in a bad state.

Note that the Plan Model can be a separate model, but it is often trained jointly with the World Model to leverage shared representations. The Plan Model can also be used independently of the state generation method used in the simulator, i.e. it can be used with a reprojective simulation or a learned World Model.

%% file: 9_reprojective_simulation.tex
\section{Reprojective Simulation}
\label{sec:reprojective_simulation}

Given a dense depth map $d_{T}$, a pose $p_{T}$, and an image $o_{T}$, we can render a new image $o^{\prime}_{T}$ by reprojecting the 3D points in the depth map to the new pose $p^{\prime}_{T}$. This process is called Reprojective Simulation. In practice, we can use a history of images and depth maps to reproject the image and inpaint the missing regions. An example is shown in Figure \ref{fig:reprojective_simulation_depth}.

\input{reprojective_simulation_depth.tex}

\subsection{Limitations of Reprojective Simulation}
\label{sub:limitations_of_reprojective_simulation}

We list some of the limitations of Reprojective Simulation:

\textbf{Assumption of a static scene:} This formulation assumes that the scene is static, and does not depend on $p_{T}$, which is usually not the case. For example, swerving towards a neighboring car might cause the driver of the neighboring car to react. We refer to this issue as the counterfactual problem.

\textbf{Depth estimation inaccuracies:} The depth map $d_{T}$ is usually noisy and inaccurate, which leads to artifacts in the reprojected image $o^{\prime}_{T}$.

\textbf{Occlusions:} Regions that are occluded in $o_{T}$ ought to be inpainted in $o^{\prime}_{T}$, which is a challenging task, and also leads to artifacts in the reprojected image.

\textbf{Reflections and lighting:} By definition, Reprojective Simulation  ignores the physics of light transport. Without ray tracing or an equivalent lighting model, reprojection can only tell where surfaces are, but not how light interacts with those surfaces from a new angle. This is a major limitation for night driving scenes, leading to noticeable lighting artifacts in the reprojected image, as shown in Figure \ref{fig:reprojective_simulation_night}.

\textbf{Limited Range:} The more $p^{\prime}_{T}$ differs from $p_{T}$, the more pronounced the artifacts become. In order to limit the artifacts, we need to limit the range of simulation to small values (typically less than 4m in translation), which is especially limiting for longitudinal motion.

\textbf{Artifacts are correlated with $p^{\prime}_{T}-p_{T}$:} The artifacts in the reprojected image are correlated with the difference between the poses $p^{\prime}_{T}$ and $p_{T}$. This correlation is exploited by the policy to predict the future action, which is not desirable. We refer to this as cheating or shortcut learning \cite{geirhos2020shortcut}.

\input{reprojective_simulation_night.tex}

%% file: reprojective_simulation_depth.tex
\begin{figure*}[ht]
    \centering
    \includegraphics[width=0.6\linewidth]{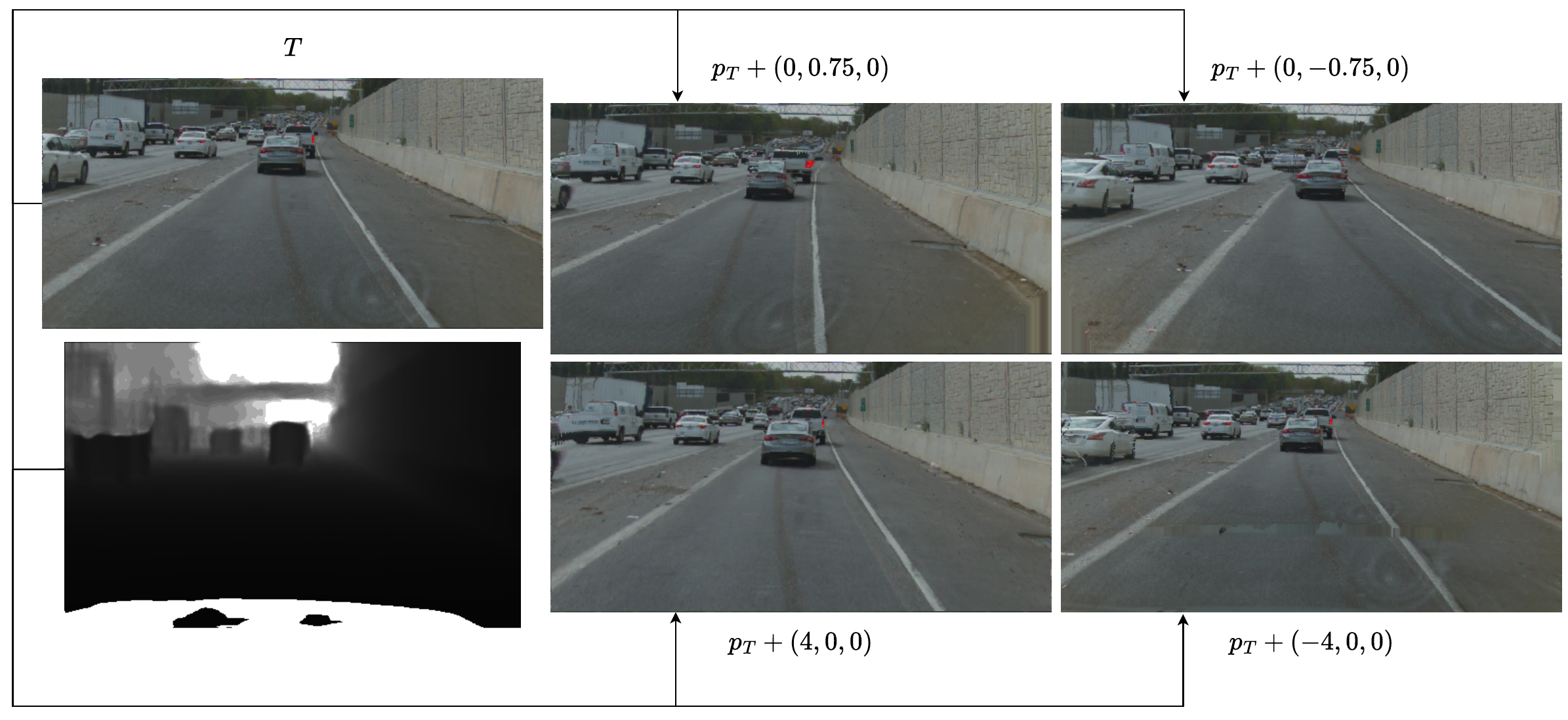}
    \caption{Left: Top: Image at $T$. Bottom: Depth map at $T$. Right: Reprojected images at $T$ using 4 different translation vectors.}
    \label{fig:reprojective_simulation_depth}
    \vspace{-0.05in}
\end{figure*}

%% file: reprojective_simulation_night.tex
\begin{figure}[ht]
    \centering
    \includegraphics[width=0.8\linewidth]{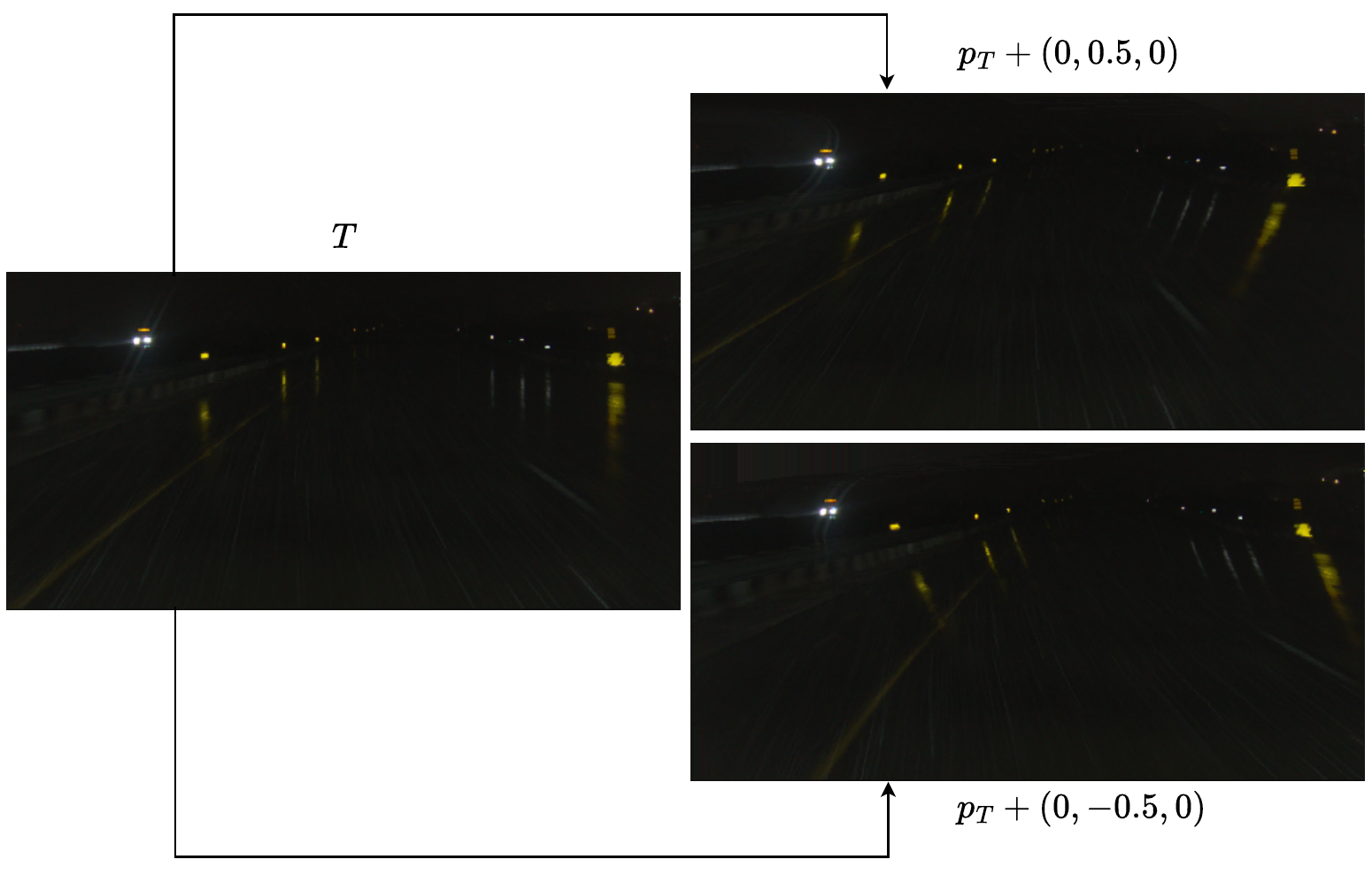}
    \caption{Left: Image at $T$, Right: Reprojected Images at $T$. Notice the lighting artifacts in the reprojected images.}
\label{fig:reprojective_simulation_night}
\vspace{-0.05in}
\end{figure}

%% file: 9_worldmodel_simulation.tex
\section{World Models Simulation}
\label{sec:world_models_simulation}

\input{world_model_simulation.tex}
The World Model simulator is a Future Anchored World Model and Plan Model.

\subsection{Video Encoder}
\label{sub:video_encoder}
We use the pretrained Stable Diffusion image VAE \cite{rombach2022high}. More specifically, \texttt{vae-ft-mse-840000-ema
-pruned} which has a compression factor of $8\times8$ and $4$ latent channels per image.
For simplicity, we exchangeably use $o$ for the latent representation of the camera image from the VAE tokenizer in this section.

\subsection{Diffusion Transformer}
\label{sub:conditional_diffusion_transformer}

\subsubsection{Architecture}
\label{subsub:architecture}

We use the Diffusion Transformer (DiT) architecture \cite{peebles2023scalable}, adapted to 3 dimensional inputs by extending the input/output patching table to a 3 dimensional table, then flattening all 3 dimensions before the Transformer blocks.

Similar to \cite{bar2024navigationworldmodels}, the vehicle poses, world timesteps, and diffusion noise timesteps are used as conditioning signals for the Diffusion Transformer. The conditioning signal embeddings are summed and passed to the Adaptive Layer Norm layer (AdaLN) \cite{xu2019understanding}. The AdaLN layer is modified to support different conditioning vectors along the time dimension.

Additionally, the attention layers use a block-wise (frame-wise) triangular causal mask, i.e. query tokens can only attend to (key, value) tokens within the same frame or the previous frames in the sequence. Note that this does not make the model physically causal, as future observations and future poses are prepended to the input sequence. However, this masking is required to use key-value caching (kv-caching) during inference, which is essential for efficient sampling.

To make it a Plan Model, the Transformer is equipped with a Plan Head, which is a stack of residual Feed Forward blocks. The Plan Head predicts the trajectory $\mathcal{T}$.

\subsubsection{Training objective}
\label{subsub:training_objective}
We adopt the Rectified Flow (RF) objective \cite{liu2022flow} for training the Conditional Diffusion Transformer. For simplicity, we omit the subscripts $T$ and $F$ for the world timestep in the following equations. We sample the noise timestep $\tau \sim \operatorname{Logit-Normal}(0.0, 1.0)$\cite{esser2024scaling} and noise the observations $o$ using Equation \ref{eq:rectified_flow_noise}.

\begin{equation}
    \label{eq:rectified_flow_noise}
    o_{\tau} = \tau \epsilon + (1 - \tau) o
\end{equation}

The Plan Head output $\mathcal{T}$ uses a Multi-hypothesis Planning loss (MHP) \cite{cui2019multimodal} with 5 hypotheses. Each hypothesis is trained using a heteroscedastic Negative Log Likelihood (NLL) loss with a Laplace prior \cite{nix1994estimating}.

The total loss $\mathcal{L}$ is a weighted sum of the Rectified Flow loss $\mathcal{L}_{\mathrm{RF}}$ and the MHP loss $\mathcal{L}_{\mathcal{T}}$ with a hyperparameter $\alpha=$ as described in Equation \ref{eq:total_loss}.

\begin{equation}
    \label{eq:total_loss}
    \begin{split}
        \mathcal{L} &= \mathcal{L}_{\mathrm{RF}} + \alpha\,\mathcal{L}_{\mathcal{T}} \\
        \text{where} \quad \mathcal{L}_{\mathrm{RF}}(o,p,\epsilon,\tau) &= \Vert w(o_{\tau}, p,\tau) - (o - \epsilon) \Vert^2 \\
        \mathcal{L}_{\mathcal{T}}(o,p,\epsilon,\tau) &= \mathrm{MHP}(w(o_{\tau}, p,\tau), \mathcal{T})
    \end{split}
\end{equation}

\subsection{Sequential Sampling}
\label{sub:sequential_sampling}
At every world timestep $T$ we use a simple Euler discretization with 15 steps $\Delta \tau = 1/15$ to sample the next latents $\tilde{o}_T$, the sampling process follows equation \ref{eq:rectified_flow_sampling}. None of the context latents $(o_{f_s},\ldots, o_{f_e}, o_1, \ldots, o_{T-1})$ are noised, and we use $\tau = 0$ as input to the model for those timesteps. The vehicle position $p_T$ can be sampled from the World Model's own Plan Head, from a Policy, from the ground truth trajectory, or artificially crafted, as described in Section \ref{sub:world_model_evaluation}.

\begin{equation}
    \label{eq:rectified_flow_sampling}
    \tilde{o}_{\tau + \Delta \tau} = \Delta \tau w(\tilde{o}_{\tau}, p,\tau + \Delta \tau) + \tilde{o}_{\tau}
\end{equation}

For the next timestep $T+1$ we shift the context latents by one timestep, and append the sampled latent $\tilde{o}_{T}$ to the context latents $(o_{f_s},\ldots, o_{f_e}, o_2, \ldots, o_{T-1}, \tilde{o}_{T})$. We repeat this process until we reach the future horizon $f_e$.

\subsection{Noise Level Augmentation}
\label{sub:noise_level_augmentation}

In order to make the model robust to the so-called "auto-regressive drift" \cite{valevski2025diffusion} i.e. errors in the Sequential Sampling process compounding frame by frame, we use a noise level augmentation technique.

For some training samples (with probability $p=0.3$), we do the following:

\begin{itemize}
    \item Sample different noise levels $\tau \sim \operatorname{Logit-Normal}(0.0, 0.25)$ at world timesteps $(1, \ldots, T-1)$,
    \item Don't noisify the future anchoring latents, i.e. $\tau = 0$ at world timesteps $(f_s, \ldots, f_e)$,
    \item Input $\tau = 0$ to the model at world timesteps $f_s, \ldots, f_e, 1, \ldots T-1$,
    \item Only compute the diffusion loss on the latents at $T$.
\end{itemize}

This diffusion noise level augmentation was essential to making the model robust to accumulated errors in the Sequential Sampling process. A similar technique was proposed in \cite{valevski2025diffusion}, although we didn't find it necessary to discretize the noise levels.

\subsection{Implementation Details and Data}
\label{sub:implementation_details_and_data}

We train three sizes of DiTs based on configurations from the GPT-2 models \cite{radford2019language}\texttt{gpt} (250M parameters), \texttt{gpt-medium} (500M parameters), and \texttt{gpt-large} (1B parameters). We use three dataset sizes: 100k, 200k, and 400k segments, each segment is 1 minute long of driving video and vehicle poses. The videos are downscaled to $128\times256$ pixels before being fed to the VAE. We downsample all the data to 5 Hz.

Data and Model size scaling results are shown in Figure \ref{fig:world_model_scaling}. Unless otherwise stated, we use the 500M DiT model trained on 400k segments for the rest of the experiments.

Every training sample is constructed as follows: we sample a context of $T=2s$  from the dataset, then we sample a world timestep $T<fs<9s$: the start of the future horizon, $f_e - f_s$ is kept constant to $1s$. The image VAE features and the vehicle poses are then concatenated as described in equation \ref{eq:future_anchored_world_model}.

At every timestep, the trajectory $\mathcal{T}$ is constructed as a sequence of positions, speeds, accelerations, orientations, and orientation rates up to a future horizon of 10 seconds.

See Figure \ref{fig:world_model_simulation} for examples of World Model rollouts.


\input{world_model_scaling.tex}

\subsection{World Model Evaluation}
\label{sub:world_model_evaluation}

We use LPIPS \cite{zhang2018unreasonable} similarity of the generated images to the ground truth images as a measure of image/video quality.

Note that the baseline LPIPS score is $\operatorname{LPIPS} = 0.148$ due to the VAE compression as measured on the test set, and the lower the LPIPS score the better the quality of the generated images.

In order to evaluate how accurately the World Model respects the vehicle positions $p_T$ inputs, we use a Pose Net to measure the error between the commanded vehicle motion and the one generated by the World Model.

The Pose Net is a supervised model trained to predict a variety of outputs, such as pose, lane lines, road edges, lead car position, etc.

\input{posenet.tex}

We evaluate the World Model over 1,500 rollouts from different segments of the test set. We can run the World Model in multiple modes.

\subsubsection{Image Quality}
\label{subsub:world_model_evaluation_image_quality}

\textbf{Observation and action teacher-forced next frame prediction:} At each world timestep $T$, we sample a frame $\tilde{o}_T$ using actions from the ground truth trajectory (rather than those from a policy). Then we replace it with the ground truth image $o_{T}$ before placing it in the context latents for the next timestep $T+1$. Results of this test are shown in left Figure \ref{fig:worldmodel_simulation_results_test_lpips}.

\subsubsection{Video Quality}
\label{subsub:world_model_evaluation_video_quality}

\textbf{Action teacher-forced sequential rollout:} Here we sample a frame $\tilde{o}_T$ using actions from the ground truth trajectory, and place it in the context latents for the next timestep $T+1$. This mode is equivalent to using the World Model normally, but with ground truth actions rather than those from a policy. Results of this test are shown in right Figure \ref{fig:worldmodel_simulation_results_test_lpips}.

\input{world_model_results_test_lpips.tex}

\subsubsection{Pose accuracy}
\label{subsub:world_model_evaluation_pose_accuracy}

\textbf{Action policy-forced sequential rollout:}
Here we sample the next frame $\tilde{o}_T$ using actions from a policy $\pi$. This is the general mode of operation for the World Model. The policy can also be a so-called "noise model". One example of a noise model is a policy that deviates from the starting position by a fixed distance as demonstrated in Figure \ref{fig:worldmodel_simulation_deviation}. Note that when the policy is the ground truth trajectory, this is equivalent to the Action teacher-forced sequential rollout.

The pose errors of the World Model measured by the Pose Net are shown in Figure \ref{fig:worldmodel_simulation_results_test_pose}.

\input{world_model_results_test_pose.tex}

We also run the World Model with the following noise model, we force a smooth lateral deviation of $\pm 0.5$m over the first 25 steps of the rollout, then let the World Model recover. We show an example of this noise in Figure \ref{fig:worldmodel_simulation_deviation}. Figure \ref{fig:worldmodel_simulation_results_test_lane_center} shows the commanded lateral deviation, and the actual deviation (measured by the Pose Net) simulated by the World Model, averaged over 1,500 rollouts. The World Model simulates the commanded deviation, but not to its full extent.

\input{world_model_simulation_deviation.tex}

\input{world_model_results_test_lane_centers.tex}

%% file: world_model_simulation.tex
\begin{figure*}[ht]
    \centering
    \includegraphics[width=0.142\linewidth]{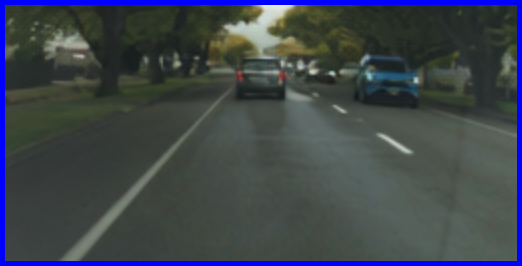}\hspace{-0.65mm}
    \includegraphics[width=0.142\linewidth]{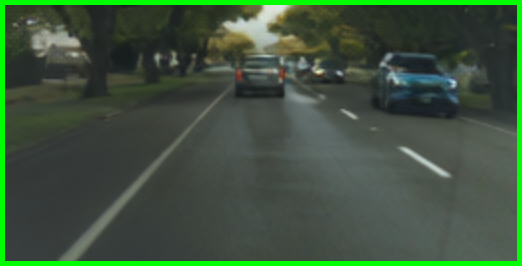}\hspace{-0.65mm}
    \includegraphics[width=0.142\linewidth]{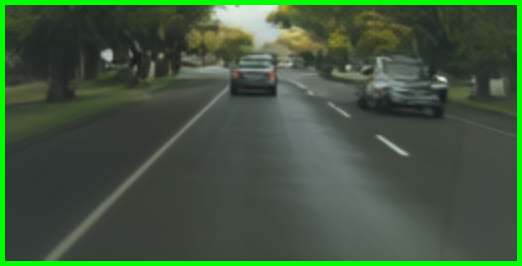}\hspace{-0.65mm}
    \includegraphics[width=0.142\linewidth]{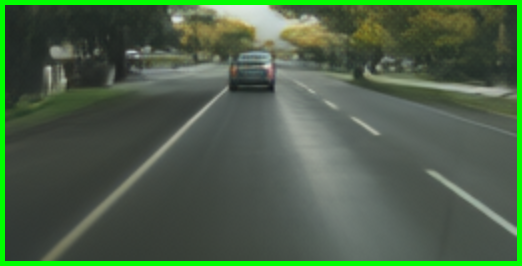}\hspace{-0.65mm}
    \includegraphics[width=0.142\linewidth]{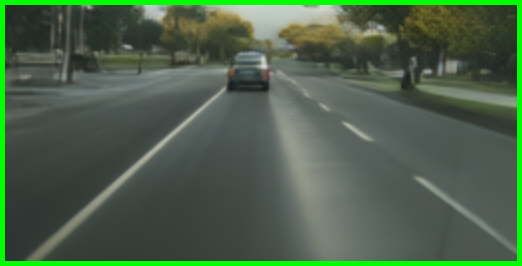}\hspace{-0.65mm}
    \includegraphics[width=0.142\linewidth]{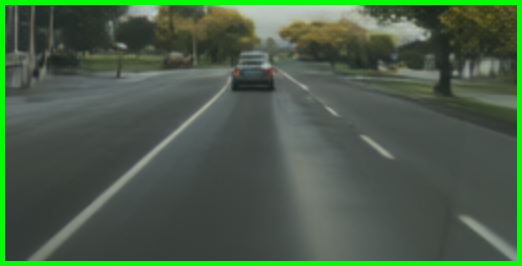}\hspace{-0.65mm}
    \includegraphics[width=0.142\linewidth]{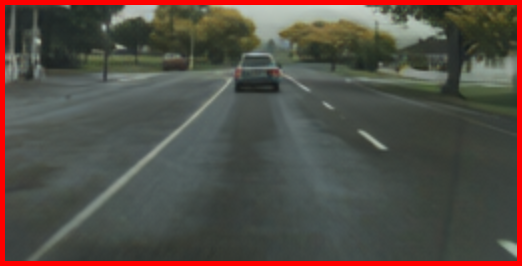}\hspace{-0.65mm} \\
    \includegraphics[width=0.142\linewidth]{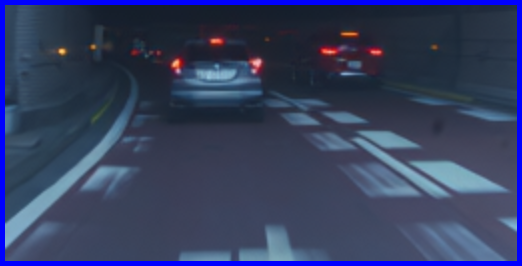}\hspace{-0.65mm}
    \includegraphics[width=0.142\linewidth]{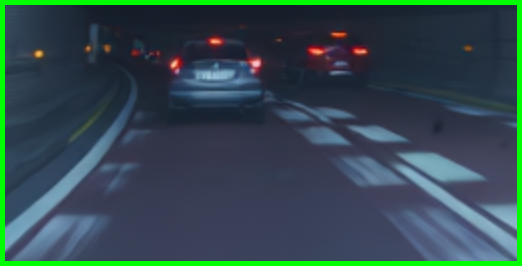}\hspace{-0.65mm}
    \includegraphics[width=0.142\linewidth]{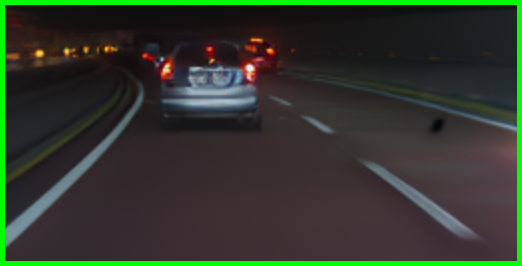}\hspace{-0.65mm}
    \includegraphics[width=0.142\linewidth]{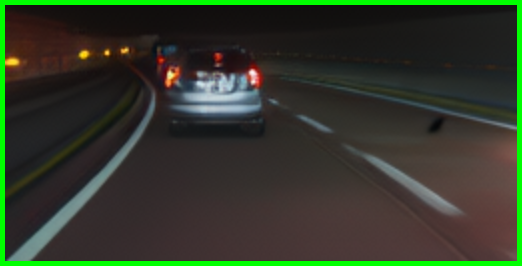}\hspace{-0.65mm}
    \includegraphics[width=0.142\linewidth]{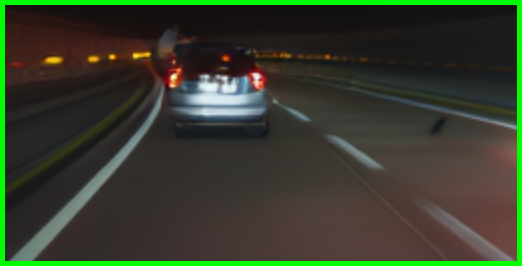}\hspace{-0.65mm}
    \includegraphics[width=0.142\linewidth]{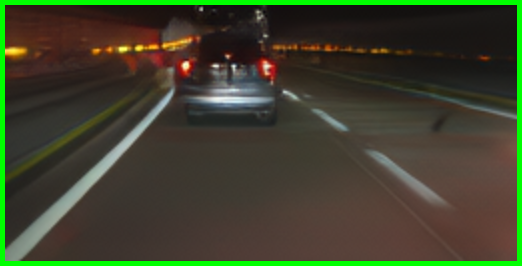}\hspace{-0.65mm}
    \includegraphics[width=0.142\linewidth]{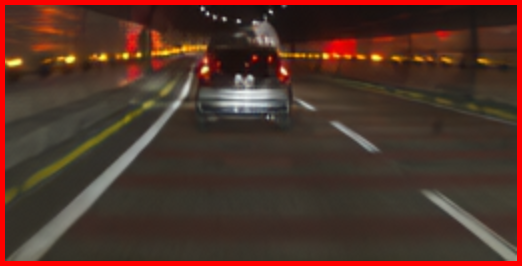}\hspace{-0.65mm} \\
    \includegraphics[width=0.142\linewidth]{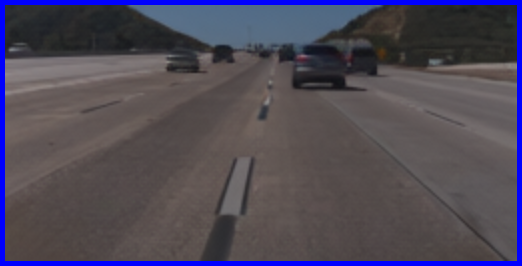}\hspace{-0.65mm}
    \includegraphics[width=0.142\linewidth]{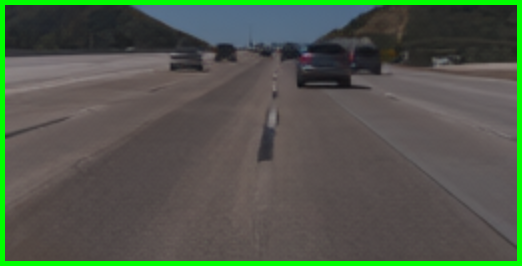}\hspace{-0.65mm}
    \includegraphics[width=0.142\linewidth]{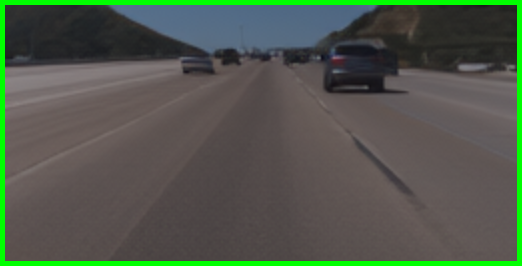}\hspace{-0.65mm}
    \includegraphics[width=0.142\linewidth]{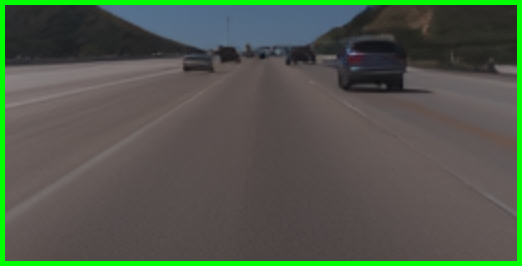}\hspace{-0.65mm}
    \includegraphics[width=0.142\linewidth]{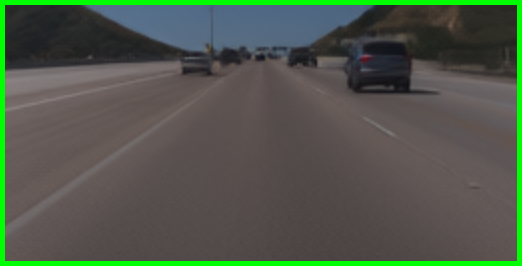}\hspace{-0.65mm}
    \includegraphics[width=0.142\linewidth]{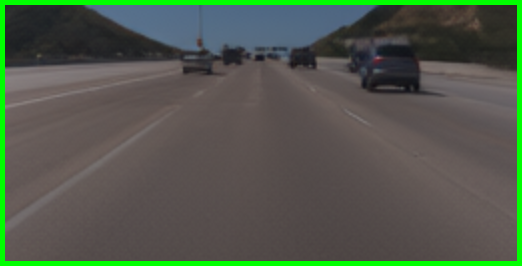}\hspace{-0.65mm}
    \includegraphics[width=0.142\linewidth]{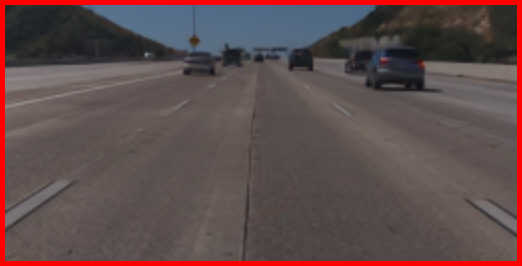}\hspace{-0.65mm} \\
    \includegraphics[width=0.142\linewidth]{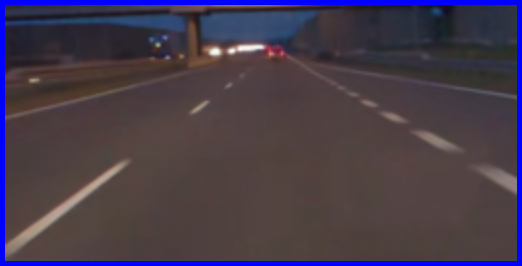}\hspace{-0.65mm}
    \includegraphics[width=0.142\linewidth]{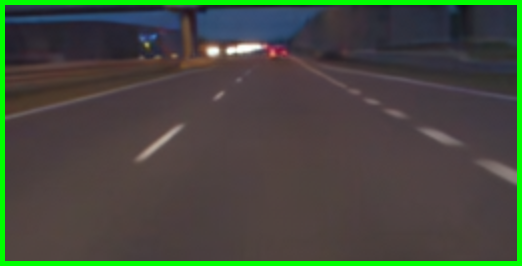}\hspace{-0.65mm}
    \includegraphics[width=0.142\linewidth]{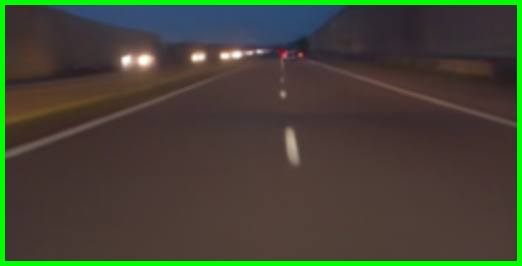}\hspace{-0.65mm}
    \includegraphics[width=0.142\linewidth]{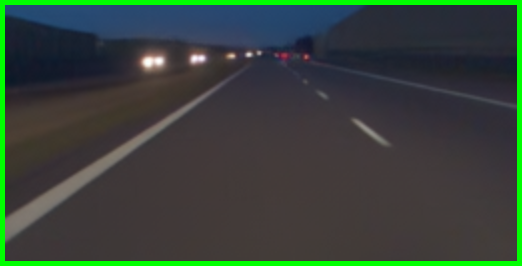}\hspace{-0.65mm}
    \includegraphics[width=0.142\linewidth]{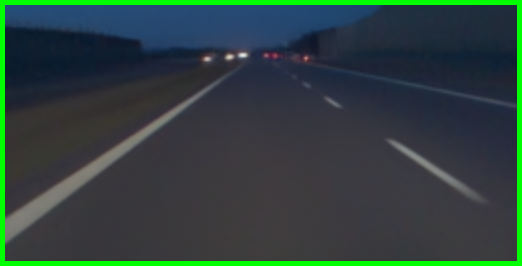}\hspace{-0.65mm}
    \includegraphics[width=0.142\linewidth]{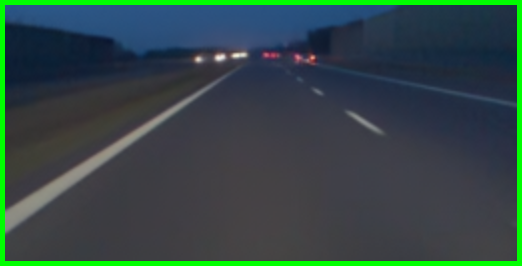}\hspace{-0.65mm}
    \includegraphics[width=0.142\linewidth]{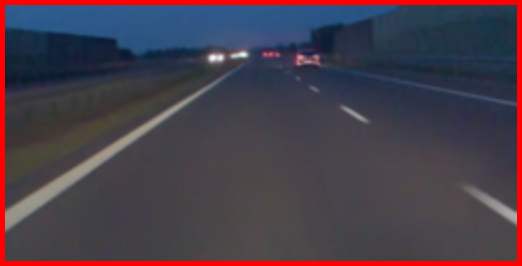}\hspace{-0.65mm} \\
    \includegraphics[width=0.142\linewidth]{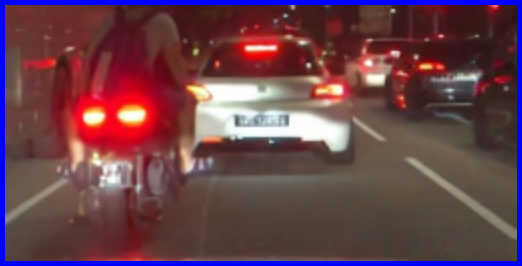}\hspace{-0.65mm}
    \includegraphics[width=0.142\linewidth]{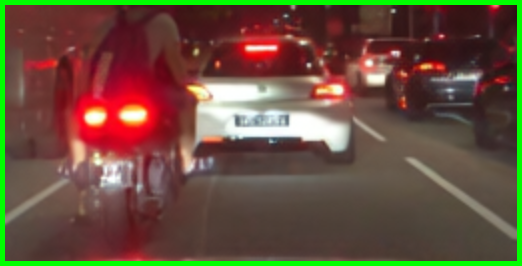}\hspace{-0.65mm}
    \includegraphics[width=0.142\linewidth]{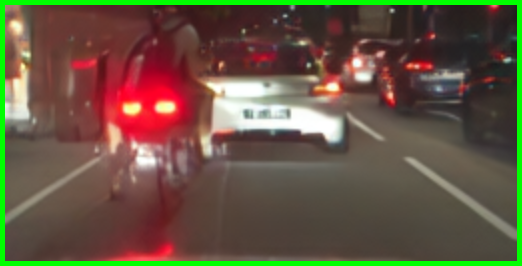}\hspace{-0.65mm}
    \includegraphics[width=0.142\linewidth]{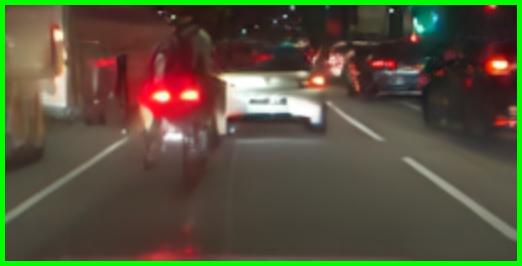}\hspace{-0.65mm}
    \includegraphics[width=0.142\linewidth]{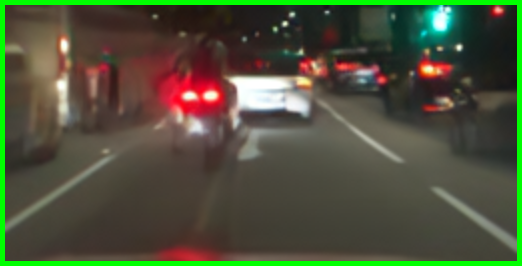}\hspace{-0.65mm}
    \includegraphics[width=0.142\linewidth]{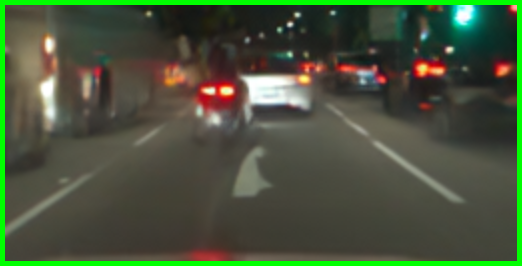}\hspace{-0.65mm}
    \includegraphics[width=0.142\linewidth]{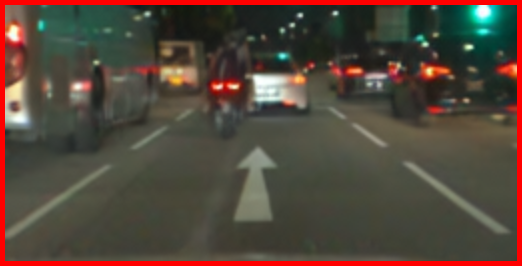}
    \caption{Five examples of World Model simulation. Blue bordered frames are the last frames of the past context, red bordered frames are the first frames of the future anchoring, and green bordered frames are simulated frames. Notice how the simulated frames comply with the future anchoring by executing lanes changes, or turning the traffic light to green.}
    \vspace{-0.05in}
    \label{fig:world_model_simulation}
\end{figure*}

%% file: world_model_scaling.tex
\begin{figure}[ht]
    \centering
    \includegraphics[width=1.\linewidth]{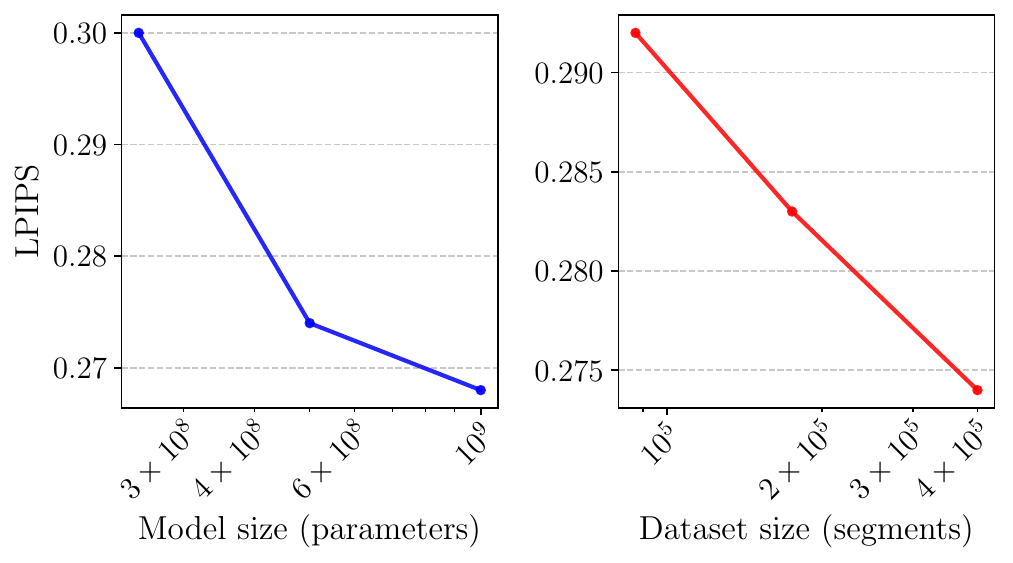}
    \caption{Left: LPIPS for different DiT model sizes, trained on 400k segments. Right: LPIPS for different dataset sizes, for a DiT of 500M parameters. Both are from the action teacher-forced sequential rollout setting.}
    \vspace{-0.22in}
    \label{fig:world_model_scaling}
\end{figure}

%% file: posenet.tex
\begin{table}[t]
    \centering
    \begin{tabular}{lcc}
    \toprule
    MAE & \makecell[c]{non VAE\\Compressed}& \makecell[c]{VAE\\Compressed} \\
    \midrule
    x speed & 0.46366 m/s & 0.59390 m/s \\
    y speed & 0.04216 m/s &  0.04393 m/s \\
    z speed & 0.04424 m/s & 0.04548 m/s \\
    roll rate & 0.00468 rad/s &  0.00524 rad/s \\
    pitch rate & 0.00433 rad/s &  0.00453 rad/s \\
    yaw rate & 0.00211 rad/s & 0.00254 rad/s \\
    y lane lines  & 0.15852 m & 0.15995 m \\
    \bottomrule
    \end{tabular}
    \caption{Pose Net MAE on non-VAE compressed and VAE compressed segments.}
    \vspace{-0.05in}
    \label{tab:pose-mae}
\end{table}

%% file: world_model_results_test_lpips.tex
\begin{figure}[ht]
    \centering
    \includegraphics[width=1.\linewidth]{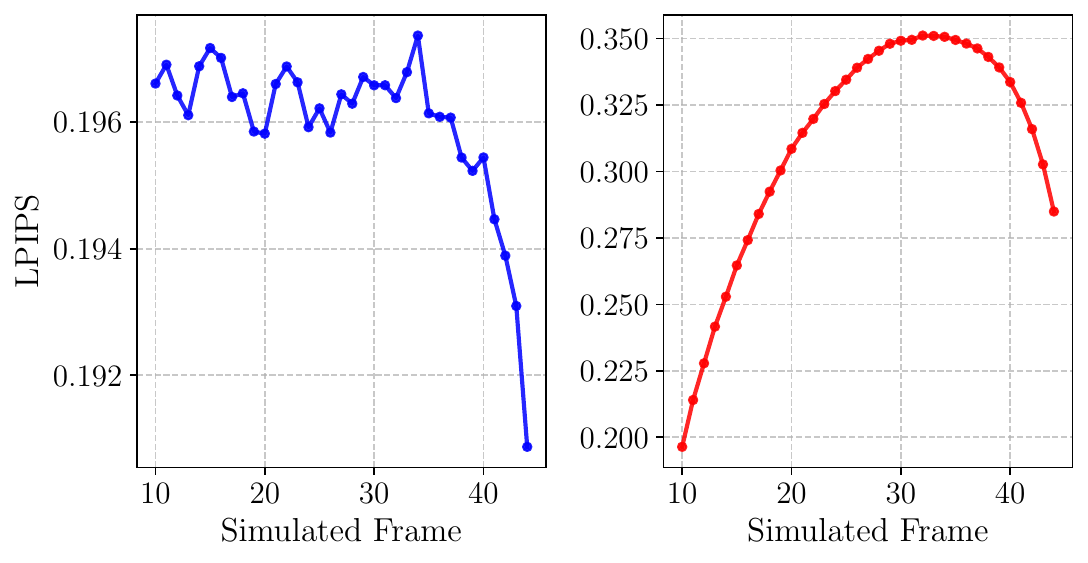}
    \caption{LPIPS for right: observation and action teacher-forced next frame prediction (image quality evaluation), left: action teacher-forced sequential rollout (video quality evaluation).}
    \label{fig:worldmodel_simulation_results_test_lpips}
    \vspace{-0.05in}
\end{figure}

%% file: world_model_results_test_pose.tex
\begin{figure}[ht]
    \centering
    \includegraphics[width=0.95\linewidth]{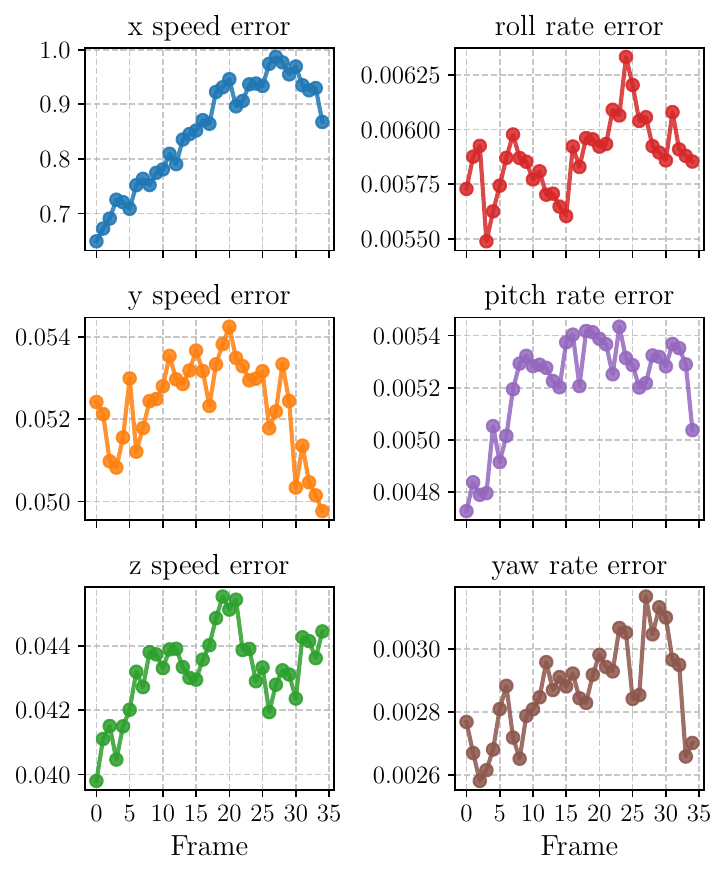}
    \caption{Action teacher-forced sequential rollout pose errors.}
    \label{fig:worldmodel_simulation_results_test_pose}
    \vspace{-0.05in}
\end{figure}

%% file: world_model_simulation_deviation.tex
\begin{figure}[htbp]
    \centering
    \begin{subfigure}[b]{0.35\textwidth}
        \centering
        \includegraphics[width=\textwidth]{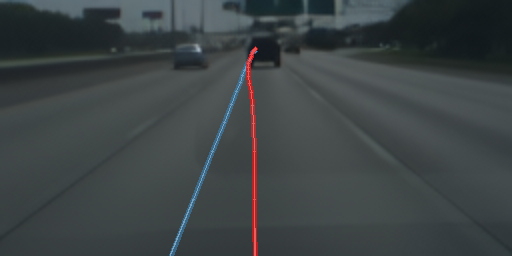}
    \end{subfigure}
    \begin{subfigure}[b]{0.35\textwidth}
        \centering
        \includegraphics[width=\textwidth]{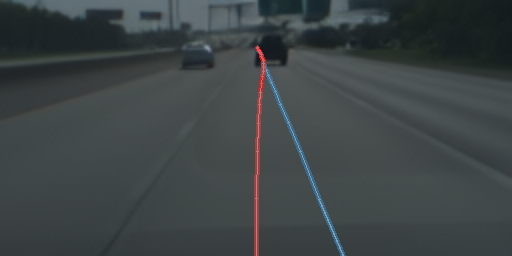}
    \end{subfigure}
    \caption{Action noise-forced sequential rollout. We force a smooth lateral deviation of $\pm 0.5$m over the 25 steps of simulation. Pictured above is the deviation to the right (top) and to the left (bottom) at step 25.}
    \vspace{-0.05in}
    \label{fig:worldmodel_simulation_deviation}
\end{figure}

%% file: world_model_results_test_lane_centers.tex
\begin{figure}[ht]
    \centering
    \includegraphics[width=0.95\linewidth]{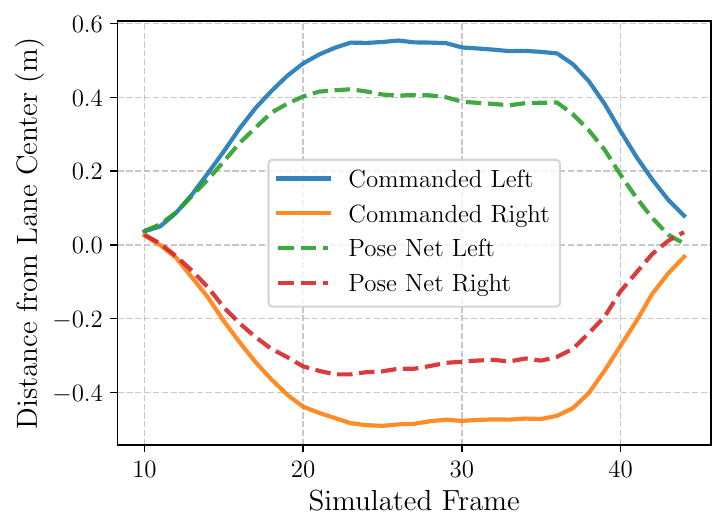}
    \caption{Deviation from the lane center in noise forced simulation. Dashed lines indicate deviation measured by the Pose Net, and solid lines indicate the commanded deviation.}
    \label{fig:worldmodel_simulation_results_test_lane_center}
    \vspace{-0.05in}
\end{figure}

%% file: 9_driving_policy_training.tex
\section{Driving Policy Training}
\label{sec:driving_policy_training}

The driving policy $\pi$ is a stack of two Neural Networks.

The first is a supervised feature extractor based on the FastViT architecture \cite{vasu2023fastvit}, which is trained to predict a variety of outputs including lane lines, road edges, lead car information, and ego car future trajectory. Note that lane lines and road edges outputs are used for visualization, and never used as part of a steering policy.

The second is a small Transformer \cite{vaswani2017attention} based temporal model predicting the same outputs as the feature extractor, in addition to the next action (desired curvature and acceleration). The temporal model's inputs are the features from the (frozen) FastViT extractor over the last 2 seconds.

Similar to the plan head of the World Model, the trajectory output of the driving policy is trained using MHP loss with 5 hypotheses and a Laplace prior. The other outputs are trained using NLL loss with a Laplace prior.

We distinguish two different approaches to training the temporal model part of the policy. Off-Policy Learning and On-Policy Learning. Off-policy learning refers to using supervised learning on the collected dataset of expert demonstrations. This is also known as imitation learning. We describe On-Policy Learning in Section \ref{sub:on_policy_learning}.

\subsection{On-Policy Learning}
\label{sub:on_policy_learning}
The temporal model is trained using a driving simulator, the policy's training samples come from its own interaction with the environment. We adopt a similar architecture to IMPALA \cite{espeholt2018impala} where a set of actors running in parallel generate experiences that are sent to a central learner. The learner updates the policy then sends a new version to the actors using a parameter server.

These experiences (also referred to as rollouts) are sequences of the form:
\begin{equation}
\begin{split}
    h^{\pi, {wp}} = &\Bigl((o_1, a_1, \hat{a}^{wp}_1), (o_2, a_2, \hat{a}^{wp}_2), \ldots, \\
    &(o_{f_s}, a_{f_s} \hat{a}^{wp}_{f_s})\Bigr),
\end{split}
\end{equation}

where $\hat{a}^{wp}$ is the action derived from the (plan equipped) Future Anchored World Model ${wp}$ during the rollout.

At every timestep $T$, the actor takes the action $a_T$ sampled from the latest policy $\pi$ available in the parameter server at the start of the rollout. The driving simulator generates the observations $o_T$ given the state of the world and the commanded action. The simulator can be ${wp}$ itself, a different World Model $w$, or any driving simulator.

The rollouts end when we reach the future horizon $f_s$, after which they are sent to the learner.

The learner optimizes the mapping $\pi: h^{\pi}_T \mapsto p(\hat{a}^w_T \mid h_T^{\pi})$, i.e. the policy learns to predict the actions that the World Model would take given the history $h_T^{\pi}$


To ensure the policy is robust to a variety of real-world effects, the Vehicle Model described in \ref{sub:vehicle_model} must model a complex distribution of action responses during training.

\subsection{Information Bottleneck}
\label{sub:information_bottleneck}
To prevent the policy from exploiting simulator-specific artifacts described in Section \ref{sub:limitations_of_reprojective_simulation}, we regularize the feature extractor by limiting the amount of information it can output to roughly 700bits. We impose this limit by adding white Gaussian noise during training, the bottleneck can be interpreted as a Gaussian communication channel with a per-sample information capacity: $\frac{1}{2} \log(1 + \text{SNR}).$ This bottleneck is similar to Gaussian Dropout \cite{rey2021gaussian} which uses multiplicative noise instead of additive noise.

\subsection{Policy Evaluation Suite}
\label{sub:policy_evaluation_suite}

We focus on evaluating the policy's lateral driving performance, i.e. its ability to accurately and smoothly steer to maintain human-like lane positioning, and successfully execute lane-changes under various conditions. Without loss of generality, longitudinal metrics and tests can be similarly integrated and are the subject of future work.

\subsubsection{Simulated On-Policy Unit Tests}
\label{sub:simulated_unit_tests}

We use the MetaDrive Simulator \cite{li2021metadrive} to evaluate the policy in closed loop. See Figure \ref{fig:metadrive} for rendered examples. We define a set of unit tests that the policy should pass:

\textbf{Convergence to lane-center test on straights and turns (24 scenarios):} We test whether the policy converges to a good position in the lane regardless of small offsets in its starting conditions. Note that the position in the lane doesn't need to be the centered, but rather a position that a human driver would converge to. See left Figure \ref{fig:policy_eval_huggig_lane_change}.

\textbf{Lane change completion test (20 scenarios):} We test whether the policy can complete a lane change maneuver regardless of small offsets in its starting conditions in the lane. During training a conditioning impulse is added prior to lane changes, and we input the same impulse during inference to trigger a lane change. See right Figure \ref{fig:policy_eval_huggig_lane_change}.

\input{metadrive.tex}

\input{policy_eval_hugging_lane_change.tex}

\subsubsection{Off Policy Evaluation}
\label{sub:off_policy_evaluation}

We use a holdout set of 1,500 segments from the dataset. We evaluate the policy on these segments off-policy by running it at each timestep and computing a variety of metrics. For simplicity, we only report the trajectory overall MAE.

\subsubsection{In the field Evaluation}
\label{sub:in_the_field_evaluation}

We evaluate the policy in the field by deploying it to openpilot \footnote{\url{https://github.com/commaai/openpilot}} and collecting data from users. openpilot is an open-source ADAS which supports a wide variety of production vehicles. The end-to-end policies described here directly control the steering actions of the vehicle to provide continuous auto-steering when the system is engaged. The longitudinal action is controlled by a classical ACC (Adaptive Cruise Control) policy, that uses lead detection and radar to slow down for other vehicles and maintains cruise speed.

\subsection{Results}
\label{sub:policy_results}

We evaluate three different policies: one trained off-policy, one trained on-policy using a reprojective simulator, and one trained on-policy using the World Model simulator. The results are shown in Table \ref{tab:metadrive_and_off_policy}. We clearly see that the policy trained off-policy fails in the on-policy tests, despite performing better in the off-policy accuracy evaluation.

Both on-policy learning methods have been successfully deployed in the real world in openpilot. Table \ref{tab:real_world} presents usage metrics collected over approximately two months of driving from a cohort of 500 users. Since openpilot is a level 2 system where disengagements are expected, we use the percentage of time and distance during which the system was engaged as the primary metric. These results demonstrate that both policies are capable of delivering meaningful driver assistance in real-world conditions.

\input{driving_policies.tex}

\input{real_world.tex}

%% file: metadrive.tex
\begin{figure}[ht]
    \centering
    \includegraphics[width=0.48\linewidth]{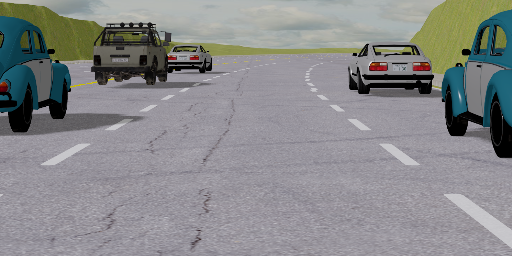}
    \includegraphics[width=0.48\linewidth]{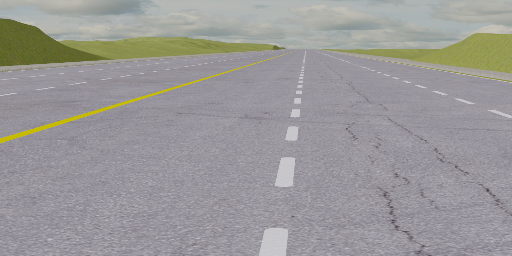}
    \caption{Examples of the MetaDrive simulated unit test scenarios.}
    \vspace{-0.05in}
    \label{fig:metadrive}
\end{figure}

%% file: policy_eval_hugging_lane_change.tex
\begin{figure}[ht]
    \centering
    \includegraphics[width=1\linewidth]{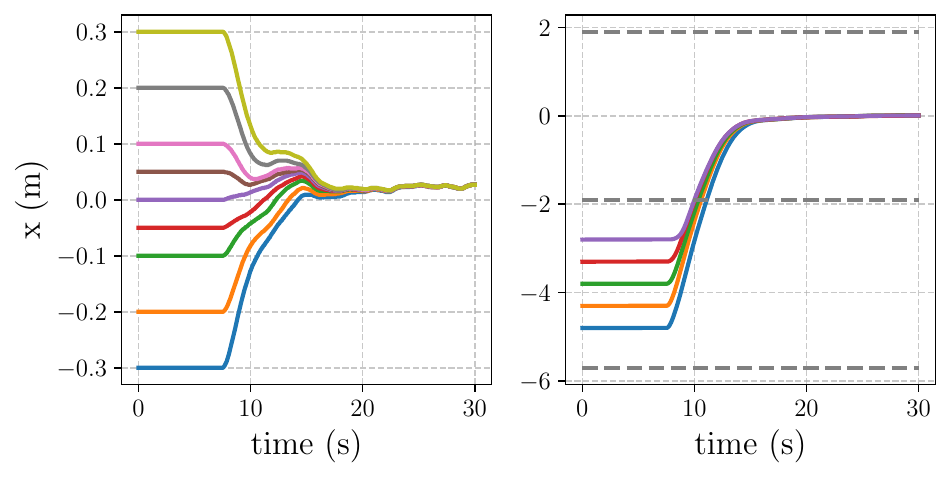}
    \caption{Left: Convergence to lane-center test results. Right: Lane change completion test results.}
    \vspace{-0.05in}
    \label{fig:policy_eval_huggig_lane_change}
\end{figure}

%% file: driving_policies.tex

\begin{table}[t]
\centering
\begin{tabular}{lccc}
\toprule
\multirow{2}{*} & Off-policy & {Reprojective}  & {World Model}  \\
\midrule
MetaDrive\\
lane center & 5/24  & 24/24  & 24/24  \\
\midrule
MetaDrive\\
lane change & 8/20 & 20/20  & 19/20 \\
\midrule
Off-policy  \\
\makecell[l]{trajectory\\MAE} & 0.361  & 0.369 & 0.394 \\
\bottomrule
\end{tabular}

\caption{Performance of driving policies trained in different conditions on the MetaDrive simulator and off-policy accuracy evaluation.}
\vspace{-0.05in}
\label{tab:metadrive_and_off_policy}
\end{table}

%% file: real_world.tex
\begin{table}[t]
    \centering
    \begin{tabular}{lcc}
    \toprule
                            &     Reprojective    &     World Model \\
    \midrule
    Number of trips              &      47,047    &     40,026      \\
    Engaged \% (time)   &      27.63\%             &      29.92\%      \\
    Engaged \% (distance)  &    48.10\%            &      52.49\%      \\
    \bottomrule
    \end{tabular}
    \caption{Performance in the field of the trained driving policies.}
    \vspace{-0.1in}
    \label{tab:real_world}
\end{table}

%% file: 9_conclusion.tex
\section{Conclusion and Future Work}

In this work, we propose an architecture for training a driving policy on-policy in a data-driven simulator based on real human driving data. This simulator produces both input images and action ground-truth based on real human driving data. We propose two different simulation strategies, one using a more traditional reprojective simulator, and another using a world model.

We show that the driving policies produced by these training strategies learn basic driving behaviors in our test suite, without any engineered behaviors during training. We also show that these policies can be used to produce useful ADAS products that assist driving in the real world.

We discuss the limitations of the reprojective simulation and how we expect the world models strategy will continue to scale with data and compute to produce better driving policies. While this work focuses on lateral driving policy, all methods generalize to longitudinal policy, in future work we expect to demonstrate useful ADAS products using end-to-end longitudinal policy trained with these methods.

\label{sec:conclusion}